# "Oddball SGD": Novelty Driven Stochastic Gradient Descent for Training Deep Neural Networks

Andrew J.R. Simpson [#1]

[#] *Centre for Vision, Speech and Signal Processing, University of Surrey*
*Surrey, UK*
[1] `Andrew.Simpson@Surrey.ac.uk`

*Abstract*—**Stochastic Gradient Descent (SGD) is arguably the most popular of the machine learning methods applied to training deep neural networks (DNN) today. It has recently been demonstrated that SGD can be statistically biased so that certain elements of the training set are learned more rapidly than others. In this article, we place SGD into a feedback loop whereby the probability of selection is proportional to error magnitude. This provides a novelty-driven *oddball SGD* process that learns more rapidly than traditional SGD by prioritising those elements of the training set with the largest novelty (error). In our DNN example, *oddball SGD* trains some 50x faster than regular SGD.**

## I. INTRODUCTION

A less-than-obvious founding assumption of stochastic gradient descent (SGD) is that the learning necessary for each element of the training set should be 'uniform'. I.e., the paths of descent should be of similar characteristics across the training set. Thus, by iteratively and randomly updating the weights for the various training examples, a stochastic search is conducted which converges on a general minimum. However, if the assumption of uniform learning does not hold, then some proportion of steps taken might be unnecessary or even counter productive. Thus, it might be useful to let the DNNs state of knowledge drive the SGD process.

During training, the error in predictions (made at the output layer) is computed with respect to the supervision data (e.g., classification data) for back propagation. This error might be interpreted as a measure of *novelty* – what the network cannot predict it does not *know,* so unknown can be interpreted as *novel* in the context of prior learning (or *forgetting* [1]). It has been demonstrated [1] that SGD may be biased towards selective learning of specific elements of a training set by frequentist statistical biases representing the probability of a given training element being selected for an SGD update step. By combining this statistical selectivity with the measure of *novelty* (prediction error), we may prioritise the most *novel* training examples for update steps during SGD.

In this article, we introduce *oddball SGD* – a novelty-driven SGD process which selectively learns those elements of the training set which are least well predicted. Using *parallel dither* [2,3] to enable non-batch SGD, we show that *oddball SGD* speeds up learning by around 50x.

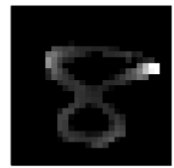

**Fig. 1. Example MNIST image.** We took the 28x28 pixel images and unpacked them into a vector of length 784 to form the input at the first layer of the DNN.

## II. METHOD

As example case, we use the well-known computer vision problem of hand-written digit classification using the MNIST dataset [4]. For the input layer we unpacked the images of 28x28 pixels into vectors of length 784. An example digit is given in Fig. 1. Pixel intensities were normalized to zero mean. Replicating Hinton's [5] architecture, but using the biased-sigmoid [6] activation function (which is optimised for demodulation), we built a fully connected network of size 784x100x10 units, where the 10-unit softmax output layer corresponds to the 10-way digit classification problem.

Operating within the so-called 'small-data regime' (as in [3]), we used only the first 256 training examples of the MNIST dataset and tested on the full 10,000 test examples. We trained several instances of the model with non-batch SGD (equivalent to a batch size of 1 in batch-averaged SGD). The first was a baseline model without regularisation. The second was the baseline model regularised with dropout. The third was the baseline model regularised with dither [2]. The fourth was the baseline model regularised with 100x parallel dither [3]. The fifth was the baseline model regularised using 100x parallel dither w/ dropout [3]. The final model was trained using *oddball SGD* and regularised with 100x parallel dither w/ dropout [3].

*Oddball SGD – novelty.* Each training iteration of *oddball SGD* began with a feed-forward pass over the 256-element training set. Absolute prediction error (the absolute difference between the prediction of the model and the training data for the output layer) was then computed, for each training element, in the output layer with respect to the training data. Then, for each element of the training set, the sum of the absolute error (across the 10-way output layer) was computed and placed in a vector (length 256) corresponding to the training examples. This vector represents the state of *novelty* of each training element.

*Novelty-driven selection statistics.* The *novelty* vector was then normalised so that it summed to 1 (i.e., it could be interpreted in terms of instantaneous probabilities). The resulting normalised *selection probability vector* was then used to assign instantaneous selection probabilities to each training element (so that selection probability was proportional to the novelty).

During each iterative step of *oddball SGD*, an element of the training set was randomly selected according to the selection probabilities. Note that this is in contrast to the traditional SGD method where the entire training set is used in random order for each full-sweep iteration.

*Parallel dither and dropout.* During non-batch SGD, each training example was replicated 100 times to form a parallel set. For parallel dither, each element of this set was dithered independently by adding uniform random noise of zero mean and unit scale and the gradients computed for each element independently. For parallel dither w/ dropout [7], both dither and dropout were applied at the same time (i.e., the parallel set was still of size: 100). Then, each parallel set of gradients (representing a single training example) was averaged and applied. Batch averaging across training examples was not applied.

Each separate instance of the model was trained for 100 full-sweep iterations of non-batch SGD (without momentum) and the test error computed (over the 10,000 test examples) at each iteration. For the *oddball SGD* model, iterations were counted cumulatively (i.e., a full sweep of SGD is 256 steps) and may be compared like for like (equal number of iterations used).

For reliable comparison, each instance of the model was trained from the exact same random starting weights. A learning rate (SGD step size) of 1 was used for all training. All dropout was at the 50% level.

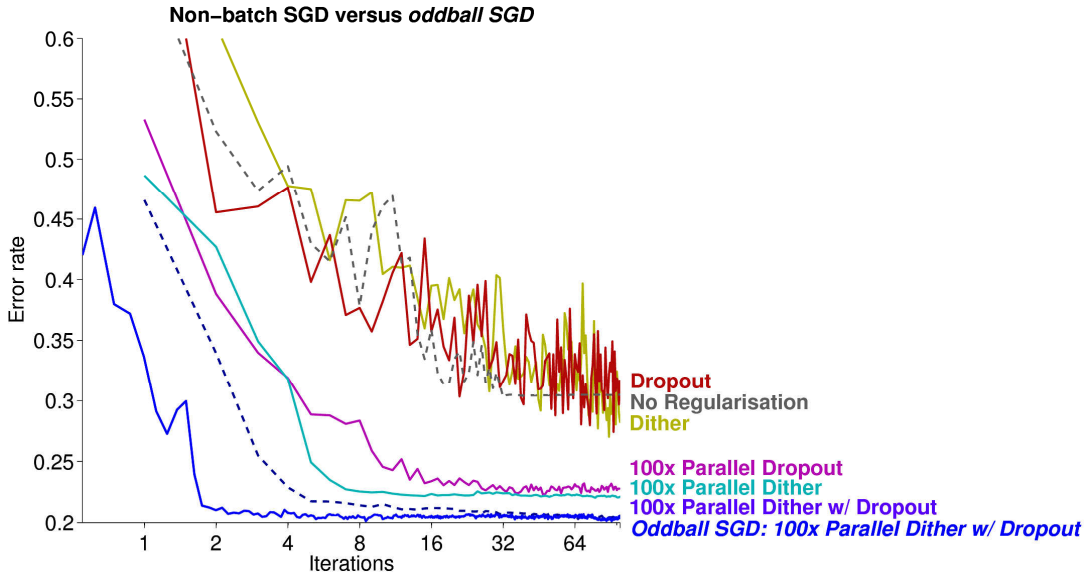

**Fig. 2. Novelty driven learning: *oddball SGD*.** Test error functions of (non-batch) SGD iterations, for the various models. All models were trained on the same data and from the same random starting weights. *NB: The x-axis is plotted on a logarithmic axis and the y-axis is somewhat cropped for better scale.*

III. RESULTS

Fig. 2 plots the test-error rates, as a function of full-sweep SGD iterations, for the various non-batch-SGD trained models and for the *oddball SGD* trained model. For convenience, the *oddball SGD* total iteration count is divided by 256, giving equivalent full-sweep cost. The *oddball SGD* model learns far more rapidly than the equivalent model trained with typical non-batch SGD. In fact, the best non-batch SGD model reaches peak performance at 100 iterations, whereas the *oddball SGD* model reaches the same performance after the equivalent of around two full-sweep iterations – i.e, it learns around 50x faster. Thus, it would seem that non-uniform novelty-based selective learning is effective.

We note, in passing, that this entire training method fails without *parallel dither* [3, see 8]. We also note that (data not shown) the same *oddball SGD* similarly improved the equivalent ReLU model (as in [8]) but the performance with ReLU was not as good as with the optimally-biased sigmoid [6] function of the present experiments [see 8 for discussion of why].

## IV. Discussion and Conclusion

In this paper we have described a novelty-driven learning algorithm – *oddball SGD* – where the running (dynamic) selection probability for each training element is proportional to the respective prediction error of the model. Thus, we have interpreted prediction error as *novelty*. Training with *oddball SGD* resulting in a speed-up of around 50x. Given that similar novelty-driven adaptation has been observed in human perception [9], it may be that a similar learning strategy is used in the brain.

## Acknowledgment

AJRS did this work on the weekends and was supported by his wife and children.

## References

[1] Simpson AJR (2015) "Use it or Lose it: Selective Memory and Forgetting in a Perpetual Learning Machine", arxiv.org abs/1509.03185.
[2] Simpson AJR (2015) "Dither is Better than Dropout for Regularising Deep Neural Networks", arxiv.org abs/1508.04826.
[3] Simpson AJR (2015) "Parallel Dither and Dropout for Regularising Deep Neural Networks", arxiv.org abs/1508.07130.
[4] LeCun Y, Bottou L, Bengio Y, Haffner P (1998) "Gradient-based learning applied to document recognition", Proc. IEEE 86: 2278–2324.
[5] Hinton GE, Osindero S, Teh Y (2006). "A fast learning algorithm for deep belief nets", Neural Computation 18: 1527–1554.
[6] Simpson AJR (2015) "Abstract Learning via Demodulation in a Deep Neural Network", arxiv.org abs/1502.04042.
[7] Hinton GE, Srivastava N, Krizhevsky A, Sutskever I, Salakhutdinov R (2012) "Improving neural networks by preventing co-adaptation of feature detectors", The Computing Research Repository (CoRR), abs/1207.0580.
[8] Simpson AJR (2015) "Taming the ReLU with Parallel Dither in a Deep Neural Network", arxiv.org abs/1509.05173.
[9] Simpson AJR, Harper NS, Reiss JD, McAlpine D (2014) "Selective Adaptation to "Oddball" Sounds by the Human Auditory System", J Neurosci 34:1963-1969.